\documentclass{article} 
\usepackage{iclr2027_conference,times}

\iclrfinalcopy

\usepackage{amsmath,amsfonts,bm}

\def\eqref#1{equation~\ref{#1}}

\def\1{\bm{1}}

\DeclareMathAlphabet{\mathsfit}{\encodingdefault}{\sfdefault}{m}{sl}
\SetMathAlphabet{\mathsfit}{bold}{\encodingdefault}{\sfdefault}{bx}{n}

\usepackage{hyperref}
\usepackage{url}

\usepackage{graphicx}
\usepackage{booktabs}   
\usepackage{graphicx}  
\usepackage{pifont}

\usepackage{amsmath}
\usepackage{booktabs}
\usepackage{algpseudocode}

\title{Evidence-Aligned Multimodal On-Policy Self-Distillation
for Fine-Grained Visual \mbox{Understanding}}

\author{
Nanxing Hu\textsuperscript{1},
Qiwei Yan\textsuperscript{2},
Jinchao Zhang\textsuperscript{*},
Guoliang Kang\textsuperscript{1,*}
\\
\textsuperscript{1}Beihang University
\\
\textsuperscript{2}University of the Chinese Academy of Sciences \\
\textsuperscript{*}Corresponding authors
}

\begin{document}

\maketitle
\fancyhead{}

\begin{abstract}
Fine-grained visual understanding requires models to recognize small details within complex images.
Multimodal on-policy self-distillation (OPSD) addresses this challenge by using a teacher conditioned on evidence-centered crops to supervise a student conditioned on original images along student-generated trajectories.
Ideally, teacher corrections, the distributional changes from the student toward the privileged teacher, should be driven by task-relevant visual evidence.
However, the designs that make the teacher effective also introduce other interference.
Using a lagged or frozen teacher improves training stability but introduces a model-state gap from the evolving student, 
while cropping enhances task-relevant evidence but also loses the visual context.
These two sources of interference make the teacher corrections not purely rely on the visual evidence. 
We introduce Evidence-Aligned multimodal on-policy self-Distillation (EAD), which retains the crop-conditioned teacher as the target but constructs a separate evidence reference for weighting the corrections.
To exclude the effect of lagged model-state from this reference, EAD measures prediction changes using the current student.
To avoid crop-induced context changes, EAD masks the evidence region in the original image while preserving the other visual context.
The change from the student's masked-image prediction to its original-image prediction provides a controlled reference for the direction in which the visual evidence shifts the student's prediction.
EAD weights each teacher correction by its cosine alignment with the reference, i.e., retaining aligned corrections and downweighting the rest.
Retaining only 6\% of the supervision mass of dense OPSD, EAD consistently outperforms previous state-of-the-art methods. 
We also show that with small-scale backbones, EAD achieves competitive performance compared to substantially larger open-weight and closed-source models.
\end{abstract}

\section{Introduction}
Fine-grained visual understanding tasks require multimodal large language models (MLLMs) to recognize small, localized evidence within large or cluttered visual contexts. 
However, reliably extracting and exploiting such evidence remains challenging, even for models with strong general-purpose visual and reasoning capabilities~\citep{khayatkhoei2025mllms,liu2025vlm,wang2026grasp}.
Recent work~\citep{yuan2026vision} addresses this limitation with On-Policy Self-Distillation (OPSD), where the student generates trajectories from the original image and a crop-conditioned teacher, with privileged access to the visual evidence, provides next-token targets conditioned on the same prefixes.
This encourages the student to make better use of fine-grained visual evidence from the original image, which has shown obvious gains on fine-grained visual understanding tasks.

The value of the teacher supervision comes from corrections driven by the task-relevant visual evidence exposed through the crop.
However, the designs that make the teacher effective also introduce other interferences into these corrections.
First, a lagged or frozen teacher is important for stable self-distillation.
Vision-OPD~\citep{yuan2026vision} shows that directly using the current student as the teacher can destabilize training. 
The trained model collapses to near-zero accuracy across all benchmarks. 
However, using a lagged or frozen teacher introduces a model-state gap, 
so part of the teacher correction can arise from differences in model state.
Second, the cropping is essential for exposing fine-grained visual evidence that may be difficult to extract from the original image, but cropping also removes surrounding visual context and changes the visual-token input presented to the model.
This will introduce crop-induced variation into the teacher correction.
Section~\ref{sec:discrepancy-profiles} quantifies these two effects at the token level, showing that both model-state and crop-induced variations substantially alter the teacher corrections.

\begin{figure*}[t]
    \centering
    \includegraphics[width=\linewidth]{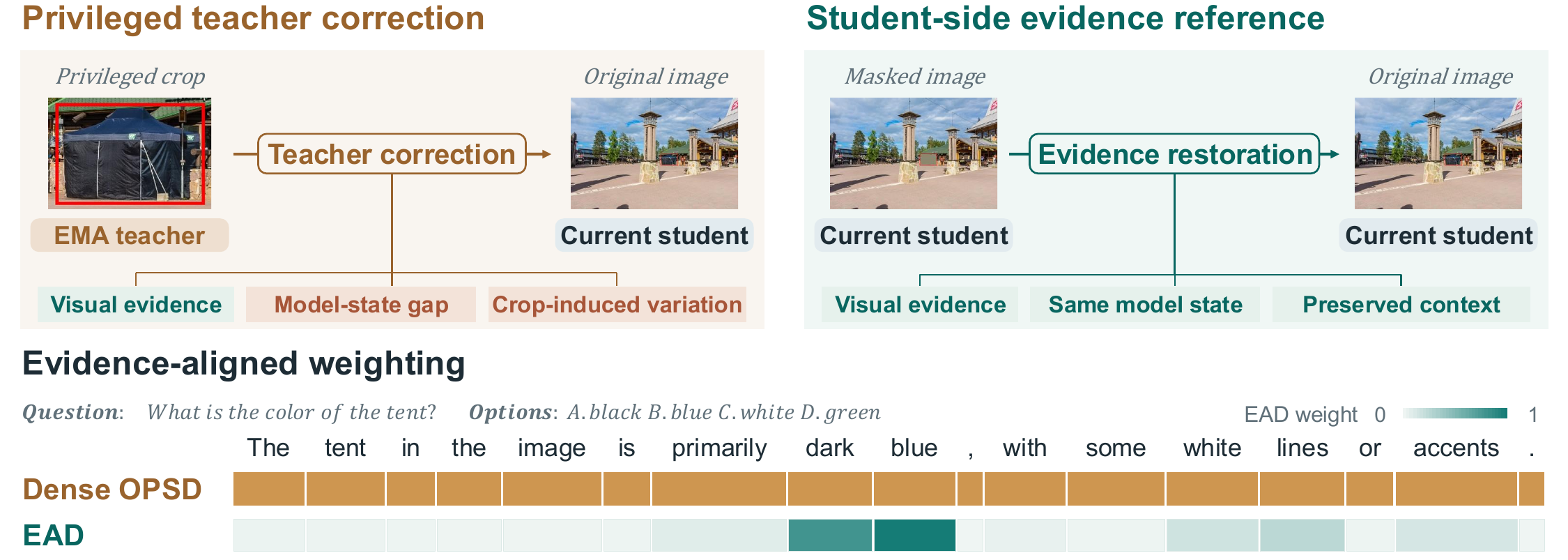}
    \caption{
    \textbf{Motivation and overview of EAD.}
    The privileged teacher correction in multimodal OPSD contains both the desired evidence-driven signal and interference arising from model-state gap and crop-induced context changes (top left).
    EAD constructs a separate student-side evidence reference from the current student's prediction shift induced by evidence restoration, which keeps the model state fixed and preserves the surrounding visual context (top right).
    EAD then weights teacher corrections by the alignment with this evidence reference, concentrating supervision on visually supported corrections (bottom).
    }
\label{fig:overview}
\end{figure*}

Therefore, we introduce Evidence-Aligned multimodal on-policy self-Distillation (EAD), as illustrated in Figure~\ref{fig:overview}.
EAD retains the crop-conditioned EMA teacher as the supervision target, while constructing a separate evidence reference to determine how much of each teacher correction to retain.
To exclude model-state gap from this reference, EAD computes the prediction change of current student.
To avoid crop-induced context changes, EAD masks only the evidence region in the original image while preserving the surrounding visual context.
Under the same generation prefix, the change from the student's masked-image prediction to its original-image prediction provides a controlled reference for the direction in which the visual evidence shifts the student's prediction.
EAD then compares each privileged teacher correction with this reference.
Their directional agreement is measured by cosine similarity and directly determines the token-level supervision weight: 
downweighting weakly aligned corrections and excluding opposing ones from supervision.

Retaining only 6\% of the supervision mass of dense OPSD, EAD consistently outperforms previous state-of-the-art methods. 
We also show that with small-scale backbones, EAD achieves competitive performance compared to substantially larger open-weight and closed-source models.
These results suggest that OPSD is most effective when teacher corrections are selectively retained according to their alignment with privileged evidence.

Our contributions are 
(1) We identify two sources of interference in multimodal OPSD: model-state gap and crop-induced context loss. Our analysis shows that they substantially affect the teacher corrections.
(2) We propose EAD, which retains the crop-conditioned EMA teacher as the supervision target and constructs a separate evidence reference from the shift in the current student's prediction caused by restoring the masked visual evidence. EAD uses the directional alignment between the teacher correction and the evidence reference to weight the distillation supervision.
(3) Experiments show that EAD consistently outperforms previous state-of-the-art methods and is competitive with substantially larger models with retaining only about 6\% of the supervision mass of dense OPSD.

\section{Related Work}

\paragraph{Fine-Grained Visual Understanding.}
Fine-grained visual understanding requires MLLMs to recover small,
localized evidence from high-resolution or cluttered images.
Existing methods often enhance fine-grained perception at inference time through visual search, localization, or image manipulation~\citep{wu2024v,khayatkhoei2025mllms,wang2025divide,zheng2026deepeyes,zhang2508thyme}.
Recently, some approaches improve fine-grained perception through training without additional test-time operations.
Zooming without Zooming~\citep{wei2026zooming} distills the benefits of zooming into full-image training, enabling single-pass fine-grained perception. 
Vision-OPD~\citep{yuan2026vision} uses crop-based teacher supervision to improve recognition of fine details in the original image.
Our work also aims to strengthen fine-grained perception without additional inference-time cropping or tool use.

\paragraph{On-Policy Self-Distillation.}
Knowledge distillation transfers prediction distributions from a teacher to a student~\citep{hinton2015distilling}, and has been extended to autoregressive sequence generation~\citep{kim2016sequence}.
Recently, on-policy distillation~\citep{gu2024minillm,agarwal2024policy} applies teacher supervision to prefixes generated by the student's own policy, reducing the mismatch between training and inference prefix distributions.
On-policy self-distillation further uses the same model as both teacher
and student, with the teacher conditioned on additional privileged
information~\citep{zhao2026self}.
In multimodal on-policy self-distillation, Vision-OPD~\citep{yuan2026vision} uses an evidence-centered crop-conditioned teacher to supervise an original-image student.
Our work follows this setting but focuses on improving the quality of the teacher corrections.

\paragraph{Evidence-Aware Selective Distillation.}
Selective distillation selectively retains or reweights teacher supervision
at the sequence or token level, rather than treating all teacher signals
equally~\citep{wang2021selective,tavor2026rethinking}.
Visual interventions provide a way to identify how model predictions depend on visual evidence ~\citep{leng2024mitigating,hu2025enhancing,gao2026thinking}.
Building on this idea, multimodal distillation methods use visual contrasts to select or weight supervision.
Recent multimodal OPD methods use visual contrasts to identify supervision that is more closely tied to visual evidence~\citep{liu2026visual,qian2026med,sun2026v}.
Concurrent works in multimodal OPSD explore evidence-aware selection of teacher supervision. OPD-V~\citep{bi2026opd} uses contrastive privileged views to select visually supported supervision, while VAD~\citep{zhang2026vad} uses teacher responses to degraded crop to identify visually relevant corrections.
Our work adapts visual-evidence extraction to the particular structure of multimodal OPSD, yielding a more controlled signal for weighting teacher corrections.
\begin{figure*}[t]
    \centering
    \includegraphics[width=\linewidth]{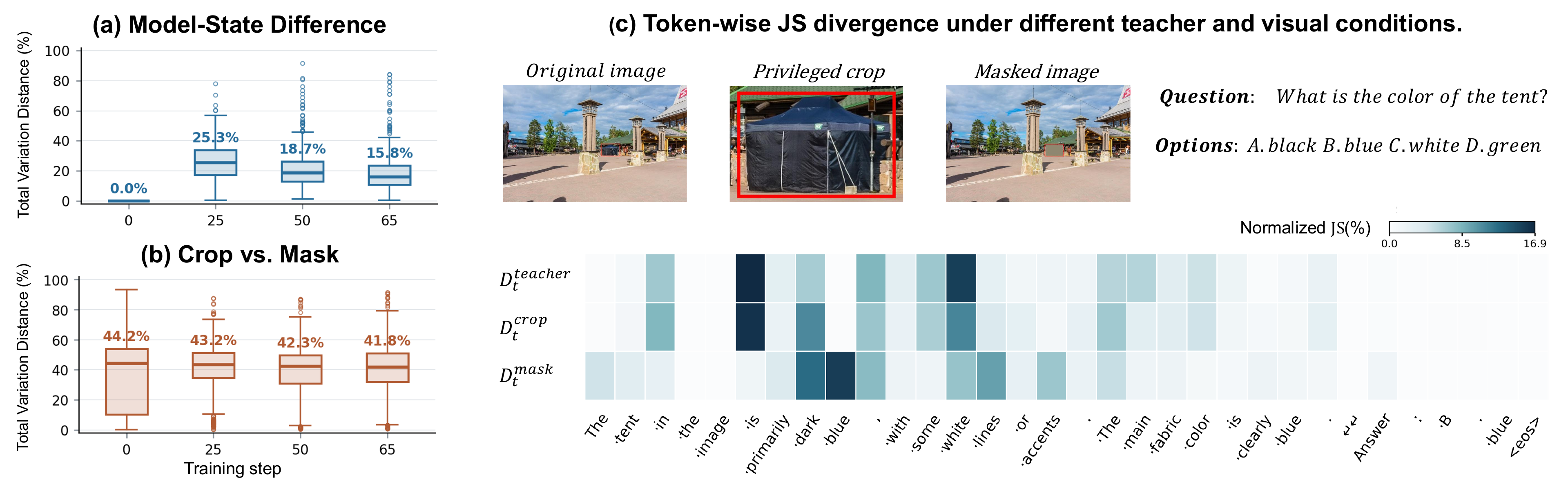}
    \caption{
    \textbf{Diagnosing interference in teacher corrections.}
    (a) Total Variation (TV) distance between the normalized token-wise JS distributions of $D_t^{\mathrm{teacher}}$ and $D_t^{\mathrm{crop}}$ over 512 images, showing the effect of model-state gap across training.
    (b) TV distance between $D_t^{\mathrm{crop}}$ and $D_t^{\mathrm{mask}}$ under the current student, showing the difference between crop-induced changes and evidence removal.
    (c) The corresponding token-wise JS distributions for one example. 
    $D_t^{\mathrm{mask}}$ places greater emphasis on tokens associated
    with the task-relevant visual evidence. $D_t^{\mathrm{crop}}$ shows large divergence on less vision-relevant tokens.
    }
    \label{fig:profiles}
\end{figure*}

\begin{figure*}[t]
    \centering
    \includegraphics[width=\linewidth]{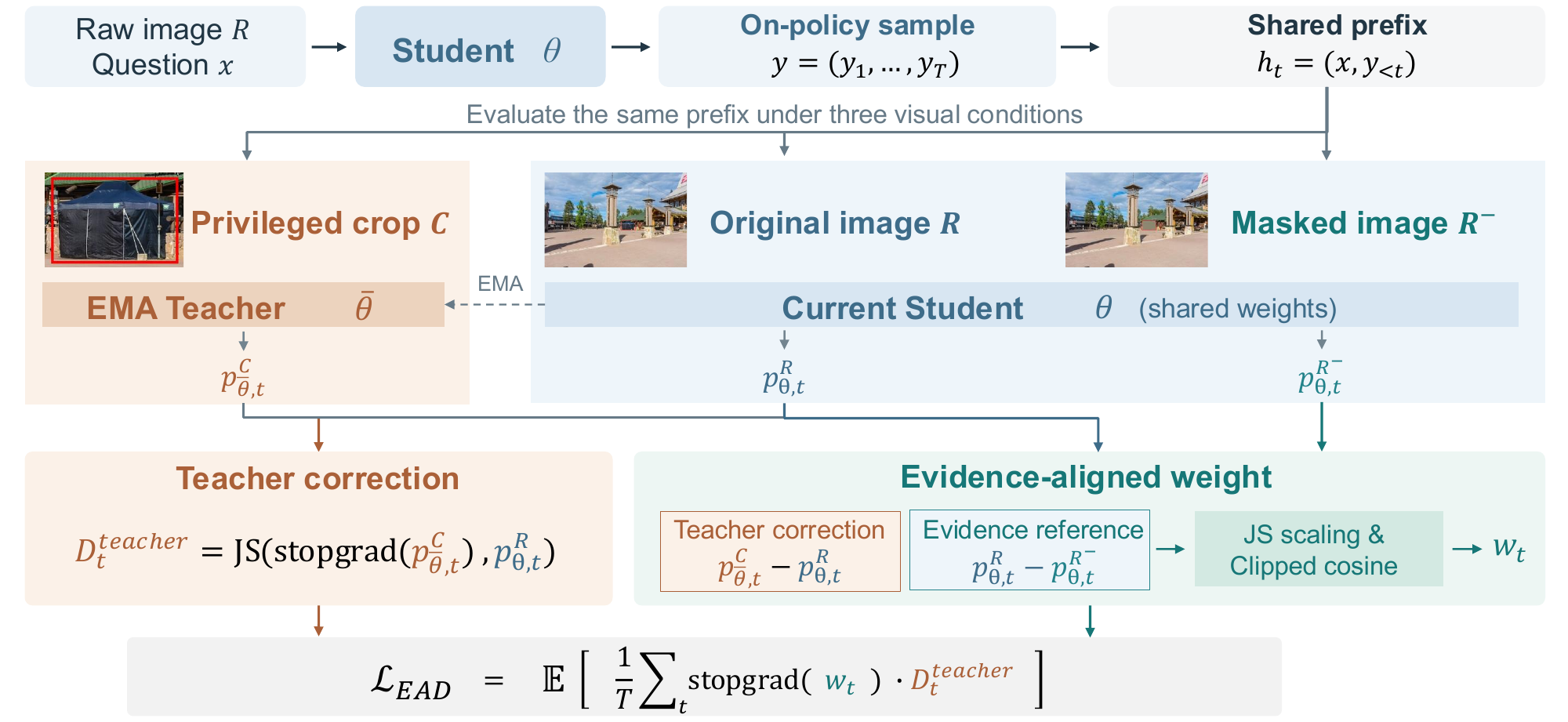}
    \caption{
    \textbf{Overview of EAD.}
    The crop-conditioned EMA teacher provides the privileged target $p_{\bar{\theta},t}^{C}$, while the current student produces $p_{\theta,t}^{R}$ and $p_{\theta,t}^{R^-}$ on the original and masked images, respectively.
    The Original--Mask prediction change $p_{\theta,t}^{R}-p_{\theta,t}^{R^-}$ forms the evidence reference. 
    EAD compares the teacher correction and evidence reference after a shared JS-motivated normalization, and uses their clipped cosine similarity to weight the distillation loss.
    }
    \label{fig:pipeline}
\end{figure*}

\section{Diagnosing Teacher Corrections}
\label{sec:discrepancy-profiles}

In Vision-OPD~\citep{yuan2026vision}, the current student parameterized by $\theta$ is conditioned on the original image $R$, while the EMA teacher parameterized by $\bar{\theta}$ is conditioned on the cropped image $C$. 
Both are evaluated under the same student-generated prefix $h_t=(x,y_{<t})$, where $x$ denotes the input prompt and $y_{<t}$ the generated response tokens.
For visual input $I$, we denote the next-token distribution as $p_{\theta,t}^{I}=p_{\theta}(\cdot\mid I,h_t)$.
The crop-conditioned teacher provides $p_{\bar{\theta},t}^{C}$ as the
target for the original-image student prediction $p_{\theta,t}^{R}$.
Let $\operatorname{sg}$ denote the stop-gradient operator. The token-level distillation divergence is
\begin{equation}
D_t^{\mathrm{teacher}}
=
\operatorname{JS}\!\left(
\operatorname{sg}(p_{\bar{\theta},t}^{C}),p_{\theta,t}^{R}
\right).
\label{eq:teacher-divergence}
\end{equation}
So, the teacher distillation divergence involves both a change in model state ($\theta$ versus $\bar{\theta}$) and visual input ($R$ versus $C$), motivating a further diagnosis of how these factors influence the correction.

\subsection{Model State difference Alters Teacher Corrections}
To isolate the effect of model state, we compare the teacher correction $D_t^{\mathrm{teacher}}$ with a single-state variation that uses the current student for both visual inputs:
\begin{equation}
D_t^{\mathrm{crop}}
=
\operatorname{JS}\!\left(
p_{\theta,t}^{R},
p_{\theta,t}^{C}
\right).
\end{equation}
To quantify how the two divergences differ across response tokens, we normalize their token-level JS values within each response and compute the Total Variation (TV) distance between the resulting distributions, with $\mathrm{TV}\in[0,1]$:
\begin{equation}
\pi_t^{k}
=
\frac{D_t^{k}}{\sum_{t'} D_{t'}^{k}},
\qquad
\operatorname{TV}
\left(
\pi^{\mathrm{teacher}},
\pi^{\mathrm{crop}}
\right)
=
\frac{1}{2}
\sum_t
\left|
\pi_t^{\mathrm{teacher}}
-
\pi_t^{\mathrm{crop}}
\right|,
\label{eq:discrepancy-profile}
\end{equation}
where $k\in\{\mathrm{teacher},\mathrm{crop}\}$.
We analyze a Qwen3.5-4B model trained with Vision-OPD~\citep{yuan2026vision} using 512 randomly sampled images across training checkpoints.
As shown in Figure~\ref{fig:profiles}(a), the two token-level distributions coincide at initialization and diverge during training.
The median TV distance reaches $25.3\%$ at step 25 and remains substantial throughout training, showing that model-state gap alters the allocation of supervision across response tokens.

\subsection{Crop-Induced Change Differs from Evidence Removal}
We next examine whether the prediction change induced by the privileged crop faithfully reflects the visual evidence.
We decompose the original image as $R=(E,B)$, where $E$ denotes the task-relevant evidence and $B$ the remaining visual context.
The Raw--Crop comparison changes both components: cropping emphasizes
$E$ while removing $B$.
To obtain a cleaner evidence reference, we construct a masked image $R^-=(E^-,B)$ which masks the evidence region while preserving the surrounding context.
Using the same current student and generation prefix, we compute
\begin{equation}
D_t^{\mathrm{mask}}
=
\operatorname{JS}\!\left(
p_{\theta,t}^{R},
p_{\theta,t}^{R^-}
\right).
\label{eq:mask-discrepancy}
\end{equation}
Following Eq.~\ref{eq:discrepancy-profile}, we normalize $D_t^{\mathrm{mask}}$ and $D_t^{\mathrm{crop}}$ within each response and compute the TV distance between them.
Figure~\ref{fig:profiles}(b) shows consistently large median TV distances of approximately $42$--$44\%$ across training checkpoints.
These results reveal a substantial token-level difference between crop-induced change and evidence removal, highlighting the influence of surrounding-context removal in the cropped view.

\subsection{$D_t^{\mathrm{mask}}$ Emphasizes Visually Relevant Tokens}
Figure~\ref{fig:profiles}(c) compares the token-wise JS distributions of the three supervision signals in a color-recognition example.
Replacing the EMA teacher with the current student redistributes JS mass across response tokens.
More importantly, $D_t^{\mathrm{mask}}$ places greater emphasis on tokens associated with the task-relevant visual evidence, such as \texttt{blue}, whereas $D_t^{\mathrm{crop}}$ shows large divergence on tokens less directly tied to the visual evidence, such as \texttt{is}.
This suggests that the Original--Mask comparison is more closely tied to task-relevant visual evidence, motivating its use as the evidence reference for weighting teacher corrections. More case studies are provided in Appendix~\ref{app:case-studies}.

\section{Evidence-Aligned multimodal on-policy self-Distillation}
\label{sec:method}

Motivated by the analysis in Section~\ref{sec:discrepancy-profiles}, we introduce Evidence-Aligned multimodal on-policy self-Distillation (EAD).
As illustrated in Figure~\ref{fig:pipeline}, EAD retains $p_{\bar{\theta},t}^{C}$ as the privileged teacher target, while using the current student's Original--Mask prediction change $p_{\theta,t}^{R}-p_{\theta,t}^{R^-}$ as a separate evidence reference.
EAD compares this reference with the teacher correction vector $p_{\bar{\theta},t}^{C}-p_{\theta,t}^{R}$ and uses their directional agreement to weight the distillation loss.

\subsection{Aligning Teacher Corrections with Visual Evidence}
\label{sec:evidence_reference}

At response step $t$, the crop-conditioned EMA teacher provides the privileged target distribution $p_{\bar{\theta},t}^{C}$ for the original-image student distribution $p_{\theta,t}^{R}$.
EAD additionally computes the current student's next-token distribution conditioned on the masked image, denoted by $p_{\theta,t}^{R^-}$.
Since $p_{\theta,t}^{R}$ and $p_{\theta,t}^{R^-}$ are produced by the same model under the same generation prefix and differ only in the evidence region, their distributional change provides a controlled reference for the prediction shift induced by the visual evidence.

For two distributions $p$ and $p+\delta$, a second-order Taylor expansion of the Jensen--Shannon divergence around $p$ gives
\begin{equation}
\operatorname{JS}(p,p+\delta)
\approx
\frac{1}{8}
\sum_j
\frac{\delta_j^2}{p_j}.
\label{eq:js_taylor}
\end{equation}
where $j$ indexes vocabulary entries.
This motivates representing distributional changes in the locally normalized coordinates $\delta/\sqrt{p}$, so that the directional
comparison is consistent with the geometry of Jensen--Shannon divergence.
Based on the shared current student distribution $p_{\theta,t}^{R}$, we define
\begin{equation}
u_t^{\mathrm{corr}}
=
\frac{
p_{\bar{\theta},t}^{C}-p_{\theta,t}^{R}
}{
\sqrt{\max(p_{\theta,t}^{R},\epsilon)}
},
\qquad
u_t^{\mathrm{evid}}
=
\frac{
p_{\theta,t}^{R}-p_{\theta,t}^{R^-}
}{
\sqrt{\max(p_{\theta,t}^{R},\epsilon)}
}.
\label{eq:ead_vectors}
\end{equation}
$u_t^{\mathrm{corr}}$ represents the direction from the student distribution toward the privileged teacher target, while $u_t^{\mathrm{evid}}$ represents the prediction shift induced by the visual evidence. $\epsilon$ is a small numerical constant.

We quantify the directional agreement using the clipped cosine similarity:
\begin{equation}
w_t
=
\max\left(
\frac{
\langle
u_t^{\mathrm{corr}},
u_t^{\mathrm{evid}}
\rangle
}{
\|u_t^{\mathrm{corr}}\|_2\,
\|u_t^{\mathrm{evid}}\|_2
},
0
\right),
\label{eq:ead_weight}
\end{equation}
Teacher corrections that agree more strongly with the evidence-induced
direction receive larger weights, while opposing directions receive
zero weight.

\subsection{Learning from Evidence-Aligned Corrections}
\label{sec:evidence_aligned_distillation}
We use $w_{i,t}$ to weight the original distillation loss, yielding the final EAD objective:
\begin{equation}
\mathcal{L}_{\mathrm{EAD}}
=
\frac{1}{B}
\sum_{i=1}^{B}
\frac{1}{T_i}
\sum_{t=1}^{T_i}
\operatorname{sg}(w_{i,t})
\operatorname{JS}\!\left(
\operatorname{sg}\!\left(p_{\bar{\theta},i,t}^{C}\right),
p_{\theta,i,t}^{R}
\right),
\label{eq:ead_loss}
\end{equation}
where $B$ is the number of on-policy responses, $T_i$ is the number of valid response tokens, and $\operatorname{sg}$ denotes stop-gradient.
The weighted token losses are averaged within each response and then equally across responses.
The teacher and masked-image branch are used only during training, and inference only uses the trained student on the original image.

\begin{table*}[t]
\centering
\small
\setlength{\tabcolsep}{5.2pt}
\renewcommand{\arraystretch}{1.08}
\caption{
\textbf{Fine-grained visual understanding.}
We report results on six fine-grained visual benchmarks under the single-forward-pass evaluation setting.
Results for closed-source and open-weight models are taken from \citet{yuan2026vision}, while all Qwen3.5-4B/9B-based methods are obtained from our evaluation.
All post-training methods use the same backbone and training data, are trained for one epoch, and are evaluated using the final checkpoint.
Avg.\ is the unweighted mean over the six benchmarks. Best results within each Qwen3.5 scale are shown in bold.
}
\label{tab:fine-grained}
\resizebox{\textwidth}{!}{%
\begin{tabular}{lcccccccc}
\toprule
\textbf{Model}
& \textbf{Params}
& \textbf{V*Bench}
& \textbf{ZoomBench}
& \multicolumn{2}{c}{\textbf{HR-Bench}}
& \multicolumn{2}{c}{\textbf{MME-RealWorld}}
& \textbf{Avg.} \\
\cmidrule(lr){5-6}
\cmidrule(lr){7-8}
& & & &
\textbf{4K}
& \textbf{8K}
& \textbf{EN}
& \textbf{CN}
& \\
\midrule

\multicolumn{9}{c}{
\textbf{Closed-Source Models (Single Forward Pass)}
} \\
\midrule
GPT-5.2~\citep{openai2025gpt52}
& -- & 79.06 & 50.89 & 81.12 & 78.38 & 72.60 & 68.80 & 71.81 \\
GPT-5.4~\citep{openai2026gpt54}
& -- & 76.96 & 52.66 & 84.00 & 77.88 & 74.20 & 70.93 & 72.77 \\
Gemini-3.1-Pro~\citep{google2026gemini31pro}
& -- & 87.96 & 61.18 & 89.63 & 86.88 & 76.53 & 73.31 & 79.25 \\
Gemini-3.5-Flash~\citep{google2026gemini35}
& -- & 89.01 & 61.42 & 89.12 & 86.62 & 75.31 & 73.97 & 79.24 \\

\midrule
\multicolumn{9}{c}{
\textbf{Open-Weight Models (Single Forward Pass)}
} \\
\midrule
DeepEyes~\citep{zheng2026deepeyes}
& 7B & 85.86 & 46.51 & 75.13 & 72.63 & 64.10 & 64.09 & 68.05 \\
Thyme~\citep{zhang2508thyme}
& 7B & 82.20 & 45.09 & 77.00 & 72.00 & 64.80 & 64.59 & 67.61 \\
DeepEyesV2~\citep{hong2026deepeyesv2}
& 7B & 81.68 & 44.97 & 77.88 & 73.75 & 64.90 & 65.07 & 68.04 \\
SenseNova-MARS~\citep{chng2025sensenova}
& 8B & 92.15 & 47.81 & 83.13 & 78.38 & 67.90 & 68.90 & 73.05 \\
MiMo-VL-RL~\citep{yue2025mimo}
& 7B & 83.25 & 45.68 & 73.50 & 69.38 & 62.73 & 55.89 & 65.07 \\
ZwZ~\citep{wei2026zooming}
& 8B & 87.96 & 56.69 & 83.63 & 81.75 & 66.57 & 68.09 & 74.12 \\
MiniCPM-V-4.5~\citep{yu2026minicpm}
& 9B & 70.68 & 42.60 & 69.63 & 61.50 & 62.65 & 61.64 & 61.45 \\
GLM-4.6V~\citep{hong2025glm}
& 106B & 86.91 & 50.06 & 82.13 & 78.88 & 65.57 & 65.62 & 71.53 \\
Qwen3-VL-Instruct~\citep{bai2025qwen3}
& 235B & 91.10 & 56.09 & 86.13 & 80.38 & 71.74 & 69.04 & 75.75 \\
Qwen3.5~\citep{qwen35}
& 397B & 87.96 & 57.16 & 89.38 & 85.50 & 74.82 & 69.82 & 77.44 \\
Kimi-K2.6~\citep{team2026kimi}
& 1T & 88.48 & 53.14 & 81.88 & 78.00 & 69.22 & 66.13 & 72.81 \\

\midrule
\multicolumn{9}{c}{
\textbf{Qwen3.5-4B Post-Training Strategies}
} \\
\midrule
Vanilla & 4B & 82.20& 48.76& 86.12& 79.62& 58.81& 60.44 & 69.33\\
GRPO~\citep{shao2024deepseekmath}
& 4B & 86.91 & 58.93 & 84.50 & 79.38 & 72.49 & 70.78 & 75.50 \\
Vision-OPD~\citep{yuan2026vision}
& 4B & 87.43 & 59.76 & 80.62 & 80.25 & 70.77 & 69.38 & 74.70 \\
OPD-V~\citep{bi2026opd}
& 4B & 87.43 & 58.93 & 83.75 & 80.12 & 70.12 & 68.80 & 74.86 \\
VAD~\citep{zhang2026vad}
& 4B & 91.10 & 60.71 & 85.50 & 81.25
& \textbf{72.75} & 70.07 & 76.90 \\
\textbf{EAD (Ours)}
& 4B
& \textbf{91.62}
& \textbf{60.83}
& \textbf{86.38}
& \textbf{82.88}
& 72.39
& \textbf{71.10}
& \textbf{77.53} \\

\midrule
\multicolumn{9}{c}{
\textbf{Qwen3.5-9B Post-Training Strategies}
} \\
\midrule
Vanilla
& 9B
& 89.01
& 53.49
& 85.38
& 81.88
& 71.83
& 67.47
& 74.84 \\

GRPO~\citep{shao2024deepseekmath}
& 9B
& 89.53
& 57.51
& 84.75
& 81.62
& 66.47
& 65.59
& 74.25 \\

Vision-OPD~\citep{yuan2026vision}
& 9B
& 89.53
& \textbf{66.04}
& 85.62
& 82.38
& 71.11
& 69.61
& 77.38 \\

OPD-V~\citep{bi2026opd}
& 9B
& 90.05
& 62.25
& 82.75
& 82.62
& 71.65
& 70.63
& 76.66 \\

VAD~\citep{zhang2026vad}
& 9B
& 91.10
& 61.42
& 87.62
& 84.12
& 72.32
& 70.07
& 77.78 \\

\textbf{EAD (Ours)}
& 9B
& \textbf{93.19}
& 62.01
& \textbf{88.62}
& \textbf{84.62}
& \textbf{72.46}
& \textbf{71.07}
& \textbf{78.66} \\

\bottomrule
\end{tabular}%
}
\end{table*}

\begin{table}[t]
\centering
\caption{
\textbf{Effect of model state on the evidence reference.}
All variants use the same Original--Mask image pair.
\emph{Cross-state} uses different model states for the two predictions,
whereas \emph{Teacher-side} and \emph{Student-side (EAD)} use the EMA
teacher and current student alone, respectively.
Avg.\ denotes the unweighted mean over the six benchmarks.
Best results in each column are shown in bold.
}
\label{tab:evidence-model}
\begin{tabular}{lccccccc}
\toprule
\textbf{Model State}
& \textbf{V*Bench}
& \textbf{ZoomBench}
& \multicolumn{2}{c}{\textbf{HR-Bench}}
& \multicolumn{2}{c}{\textbf{MME-RealWorld}}
& \textbf{Avg.} \\
\cmidrule(lr){4-5}
\cmidrule(lr){6-7}
& & &
\textbf{4K}
& \textbf{8K}
& \textbf{EN}
& \textbf{CN}
& \\
\midrule

Cross-state
& 89.53
& 59.05
& 85.12
& 81.38
& 71.05
& 69.75
& 75.98 \\

Teacher-side
& 89.01
& 59.76
& 85.25
& \textbf{83.62}
& \textbf{73.18}
& \textbf{71.35}
& 77.03 \\

\textbf{Student-side (EAD)}
& \textbf{91.62}
& \textbf{60.83}
& \textbf{86.38}
& 82.88
& 72.39
& 71.10
& \textbf{77.53} \\

\bottomrule
\end{tabular}
\end{table}

\begin{table}[t]
\centering
\small
\setlength{\tabcolsep}{4pt}
\renewcommand{\arraystretch}{1.08}
\caption{
\textbf{Effect of visual contrast on the evidence reference.}
The first two variants use the current student and differ only in the visual inputs, while the VAD-style variant uses the EMA teacher on the cropped image and its downsampled counterpart.
All variants use the same EAD objective. Avg.\ denotes the unweighted mean over the benchmarks. 
Best results in each column are shown in bold.
}
\label{tab:evidence-visual}
\begin{tabular}{lccccccc}
\toprule
\textbf{Visual Contrast}
& \textbf{V*Bench}
& \textbf{ZoomBench}
& \multicolumn{2}{c}{\textbf{HR-Bench}}
& \multicolumn{2}{c}{\textbf{MME-RealWorld}}
& \textbf{Avg.} \\
\cmidrule(lr){4-5}
\cmidrule(lr){6-7}
& & &
\textbf{4K}
& \textbf{8K}
& \textbf{EN}
& \textbf{CN}
& \\
\midrule

Original--Crop
& 88.48
& 60.00
& 83.00
& 80.25
& 70.61
& 68.62
& 75.16 \\

\textbf{Original--Mask (EAD)}
& \textbf{91.62}
& \textbf{60.83}
& 86.38
& 82.88
& \textbf{72.39}
& \textbf{71.10}
& \textbf{77.53} \\

\midrule

Crop--Downsampled Crop (VAD)
& \textbf{91.62}
& 58.82
& \textbf{86.50}
& \textbf{83.00}
& 69.33
& 68.73
& 76.33 \\

\bottomrule
\end{tabular}
\end{table}

\begin{table}[t]
\centering
\small
\setlength{\tabcolsep}{4pt}
\renewcommand{\arraystretch}{1.08}
\caption{
\textbf{Effect of local JS scaling and clipped cosine similarity.}
\emph{No JS Scaling} removes the local normalization before computing directional alignment, while \emph{Shifted-Cosine} retains negatively aligned teacher corrections.
}
\label{tab:alignment-ablation}
\begin{tabular}{lccccccc}
\toprule
\textbf{Variant}
& \textbf{V*Bench}
& \textbf{ZoomBench}
& \textbf{HRBench-4K}
& \textbf{HRBench-8K}
& \textbf{MME-EN}
& \textbf{MME-CN}
& \textbf{Avg.} \\
\midrule
No JS Scaling
& 89.01 & 59.29 & 85.12 & 82.38 & 72.37 & \textbf{71.29} & 76.58 \\
Shifted-Cosine
& 90.58 & \textbf{61.18} & 83.50 & 81.25 & 72.22 & 70.05 & 76.46 \\
\textbf{EAD}
& \textbf{91.62} & 60.83 & \textbf{86.38} & \textbf{82.88}
& \textbf{72.39} & 71.10 & \textbf{77.53} \\
\bottomrule
\end{tabular}
\end{table}

\section{Experiments}
\label{sec:experiments}

\subsection{Experimental Setup}

\paragraph{Implementation.}
We closely follow the Vision-OPD~\citep{yuan2026vision} training configuration.
We train Qwen3.5-4B and Qwen3.5-9B on the same 6.2K fine-grained examples for one epoch and evaluate the final checkpoint.
Distillation uses Jensen--Shannon divergence computed from the top-100 logits with partial softmax, and the privileged teacher is updated by EMA with a rate of $0.05$.
To reduce premature collapse toward short responses that directly output the option letter, we fix the first generated token to \textit{The}, exclude this token from the training loss, and sample all subsequent tokens normally.
Further implementation details are provided in Appendix~\ref{app:ead_details}.

\paragraph{Benchmarks.}
We evaluate fine-grained visual perception on six benchmarks. 
V*Bench~\citep{wu2024v} evaluates the ability to locate and recognize fine visual details in high-resolution and visually crowded images.
ZoomBench~\citep{wei2026zooming} focuses on fine-grained perception where decisive evidence is localized and difficult to recover from the original image. 
HR-Bench~\citep{wang2025divide} evaluates fine-grained single- and cross-instance perception at both 4K and 8K resolutions.
MME-RealWorld~\citep{zhang2025mme} further evaluates high-resolution perception and reasoning in challenging real-world scenarios. We use both its English and Chinese subsets.
We additionally evaluate held-out generalization in Appendix~\ref{app:generalization}.
For evaluation, we follow the released Vision-OPD pipeline with a calibrated multiple-choice answer-extraction procedure, applied consistently to all methods. Unresolved responses are evaluated by Qwen3.5-122B-A10B~\citep{qwen35} as a fallback judge. More details are provided in Appendix~\ref{app:evaluation}.

\paragraph{Baselines.}
We compare against two groups of baselines.
First, we include representative closed-source and open-weight MLLMs under single-forward-pass inference. Their benchmark scores are taken directly from the evaluation reported by Vision-OPD.
Second, we compare alternative post-training strategies using the same Qwen3.5~\citep{qwen35} backbones and training data.
At both the 4B and 9B scales, we compare EAD against the Vanilla model, GRPO~\citep{shao2024deepseekmath}, Vision-OPD~\citep{yuan2026vision}, and the concurrent methods OPD-V~\citep{bi2026opd} and VAD~\citep{zhang2026vad}.
We retrain each post-training method on the same backbone and training data and evaluate all resulting checkpoints with the same pipeline.
All variants are trained for one epoch, and the final checkpoint is used for evaluation.



\subsection{Comparison with State-of-the-Art MLLMs}
\label{sec:fine-grained-results}
Table~\ref{tab:fine-grained} reports the comparison on six fine-grained visual understanding benchmarks.
For the Qwen3.5 post-training setting, all methods use the same backbone, training data, and evaluation pipeline. EAD achieves the highest average performance at both the 4B and 9B scales.
Compared with Vision-OPD, EAD improves the average by $2.83$ and $1.28$ points at 4B and 9B, respectively.
It also outperforms the strongest concurrent baseline, VAD, by $0.63$ and $0.88$ points at the two scales.
We further compare EAD with representative open-weight and closed-source MLLMs.
Despite using substantially smaller backbones, EAD reaches performance comparable to substantially larger models.
These consistent gains demonstrate the effectiveness of selectively retaining teacher corrections according to their alignment with visual evidence.

\begin{figure*}[t]
    \centering
    \includegraphics[width=\linewidth]{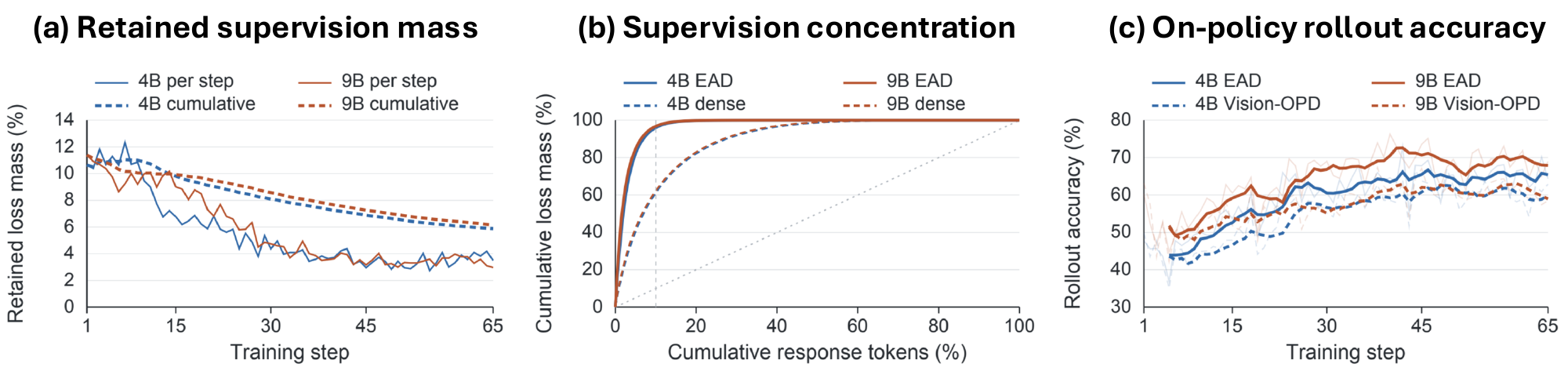}
    \caption{
    \textbf{Sparsity and concentration of EAD supervision.}
    (a) Fraction of the original distillation-loss mass retained by EAD throughout training.
    (b) Concentration of the retained loss across response tokens, compared with the original dense distillation loss on the same responses.
    (c) On-policy rollout accuracy of EAD and Vision-OPD during training. Results are shown for Qwen3.5-4B and Qwen3.5-9B.
    }
    \label{fig:supervision-analysis}
\end{figure*}

\subsection{EAD Retains Sparse and Concentrated Supervision}
\label{sec:retained-supervision}

EAD substantially reduces the amount of teacher supervision used during training, retaining only $5.87\%$ and $6.15\%$ of the original
distillation-loss mass for the 4B and 9B models, respectively (Figure~\ref{fig:supervision-analysis}(a)).
Furthermore, this reduction is not uniform: the top $10\%$ highest-loss tokens account for over $95\%$ of the retained loss mass, compared with about $61$--$62\%$ under dense distillation (Figure~\ref{fig:supervision-analysis}(b)).
Thus, EAD concentrates supervision on a much smaller subset of corrections rather than simply reducing the overall distillation strength.
Despite retaining only a small fraction of the original supervision, EAD maintains higher on-policy rollout accuracy than Vision-OPD throughout training (Figure~\ref{fig:supervision-analysis}(c)).
These results show that EAD produces a sparser and more concentrated distillation objective while maintaining more effective optimization.

\subsection{Ablation Studies}
\label{sec:evidence-reference-ablation}

We ablate the evidence-reference construction by only varying the prediction difference $\Delta_t$ used to form $u_t^{\mathrm{evid}}
=\Delta_t/\sqrt{\max(p_{\theta,t}^{R},\epsilon)}$, keeping the remaining EAD objective and training procedure fixed.
Tables~\ref{tab:evidence-model} and~\ref{tab:evidence-visual} report the fine-grained results; held-out results are provided in Appendix~\ref{app:evidence-reference-generalization}. All the experiments are conducted on Qwen3.5-4B.

\paragraph{Effect of model state.}
We first examine which model state should be used to construct the $u_t^{\mathrm{evid}}$.
All three variants use the same Original--Mask image pair and differ only in the model states used to compute the two predictions.
\emph{Cross-state} uses $\Delta_t=p_{\theta,t}^{R}-p_{\bar{\theta},t}^{R^-}$,
while \emph{Teacher-side} and \emph{Student-side (EAD)} use $\Delta_t=p_{\bar{\theta},t}^{R}-p_{\bar{\theta},t}^{R^-}$
and
$\Delta_t=p_{\theta,t}^{R}-p_{\theta,t}^{R^-}$, respectively.
As shown in Table~\ref{tab:evidence-model}, \emph{Student-side (EAD)} performs best, followed by \emph{Teacher-side}, while \emph{Cross-state} performs worst.
Both same-state variants outperform \emph{Cross-state}, showing the benefit of excluding model-state gap when constructing the evidence reference.
The further advantage of \emph{Student-side (EAD)} over \emph{Teacher-side} shows that the student's own Original--Mask prediction change provides a more effective evidence reference than the teacher's.

\paragraph{Effect of the visual contrast.} 
We next examine which image pair should be used to construct the evidence reference.
For a controlled comparison, both variants use the current student to make predictions and differ only in the visual inputs.
\emph{Original--Crop} uses $\Delta_t=p_{\theta,t}^{C}-p_{\theta,t}^{R}$,
whereas \emph{Original--Mask (EAD)} uses $\Delta_t=p_{\theta,t}^{R}-p_{\theta,t}^{R^-}$. 
As shown in Table~\ref{tab:evidence-visual}, \emph{Original--Mask (EAD)} consistently outperforms \emph{Original--Crop} across all six benchmarks.
This shows that removing the evidence while preserving the surrounding visual context provides a more effective reference than altering the context through cropping.

\paragraph{Comparison with a VAD-style reference.}
We further consider the image pair used by VAD~\citep{zhang2026vad}. 
\emph{Crop--Downsampled Crop (VAD)} uses $\Delta_t=p_{\bar{\theta},t}^{C}-p_{\bar{\theta},t}^{C_{\mathrm{deg}}}$,
where $C_{\mathrm{deg}}$ is a downsampled version of the cropped image, while keeping the EAD weighting objective and training procedure unchanged.
The \emph{Crop--Downsampled Crop (VAD)} comparison keeps both the model state and cropped region fixed, varying only the visual degradation.
However, this distributional change captures the EMA teacher's evidence sensitivity in the cropped view, rather than the current student's evidence sensitivity in the original image.
\emph{Original--Mask (EAD)} instead captures the current student's response to visual evidence within the original-image context.
As shown in Table~\ref{tab:evidence-visual}, \emph{Original--Mask (EAD)} outperforms this VAD-style reference, supporting the use of an evidence reference constructed from the current
student in the original-image context.

\paragraph{Effect of local JS scaling.}
We next examine the effect of the local JS scaling used to compare the teacher correction with the evidence reference.
\emph{No JS Scaling} removes the shared $1/\sqrt{\max(p_{\theta,t}^{R},\epsilon)}$ scaling and directly computes cosine similarity between the probability differences $p_{\bar{\theta},t}^{C}-p_{\theta,t}^{R}$ and $p_{\theta,t}^{R}-p_{\theta,t}^{R^-}$, while keeping the remaining EAD objective and training procedure unchanged.
As shown in Table~\ref{tab:alignment-ablation}, removing the local JS scaling reduces the average performance from 77.53 to 76.58.
This shows that comparing the two prediction changes under the shared local JS scaling provides a more effective alignment signal.

\paragraph{Effect of clipped cosine similarity.}
We further examine whether negatively aligned teacher corrections should be retained for supervision.
\emph{Shifted-Cosine} replaces the EAD weight
$\max(\cos(u_t^{\mathrm{corr}},u_t^{\mathrm{evid}}),0)$ with $(\cos(u_t^{\mathrm{corr}},u_t^{\mathrm{evid}})+1)/2$,
so that negatively aligned corrections retain non-zero supervision weights.
All other settings remain unchanged.
As shown in Table~\ref{tab:alignment-ablation}, \emph{Shifted-Cosine} reduces the average performance from 77.53 to 76.46.
This supports excluding teacher corrections that are directionally opposed to the evidence reference.
\section{Conclusion}

We introduced Evidence-Aligned multimodal on-policy self-Distillation (EAD) for fine-grained visual understanding.
EAD retains the privileged teacher as the supervision target while using the current student's Original--Mask prediction change as a separate evidence reference for selectively weighting teacher corrections.
By keeping the model state fixed and preserving the surrounding visual context, this reference more cleanly captures the prediction shift induced by task-relevant visual evidence.
Across 4B and 9B models, EAD consistently outperforms strong post-training baselines and remains competitive with substantially larger open-weight and closed-source models on fine-grained visual understanding benchmarks.

\subsection*{AI use statement}

Generative AI tools were used to assist with language editing and to improve the clarity, grammar, and readability of the manuscript. They were not used to formulate research hypotheses, design the methodology or experiments, analyze data, interpret experimental results, or generate scientific claims.
All AI-assisted edits were carefully reviewed and revised by the authors. 
The authors take full responsibility for the final content of the manuscript.

\bibliography{iclr2027_conference}
\bibliographystyle{iclr2027_conference}

\appendix
\section{Appendix}

\subsection{Additional Implementation Details}
\label{app:ead_details}

\paragraph{Training configuration.}
We build EAD on the Vision-OPD training pipeline and keep its training configuration unchanged unless otherwise specified.
Both Qwen3.5-4B and Qwen3.5-9B are trained for one epoch on the same 6.2K fine-grained examples used by Vision-OPD.
We use a training batch size of 96, sample $n=8$ on-policy responses per example, and optimize the student with a learning rate of $2\times10^{-6}$ and 10 warmup steps. The maximum prompt and response lengths are 8192 and 1024 tokens, respectively. 
The privileged teacher is updated by exponential moving average (EMA) with an update rate of 0.05.

\paragraph{On-policy rollout generation.}
Training trajectories are generated by the current original-image student. For each sampled response $y=(y_1,\ldots,y_T)$, all subsequent teacher and evidence reference predictions are evaluated under the same student-generated textual prefix $h_t=(x,y_{<t})$.
We fix the first generated token to \textit{The} and exclude this token from the training loss; all subsequent response tokens are sampled normally.
This construction ensures that the privileged teacher correction and the evidence reference are compared at exactly the same autoregressive states.

\paragraph{Privileged crop and evidence-masked view.}
We directly use the original-image/crop pairs released by Vision-OPD.
Thus, the crop-conditioned EMA teacher receives the same privileged visual input as Vision-OPD.
EAD introduces only an additional evidence-masked view of the original image.
We construct $R^{-}$ by masking the GT bounding-box interior with the mean RGB color of its surrounding region, while preserving the red-box annotation and the remaining image context unchanged.

\paragraph{Shared top-$K$ vocabulary support.}
Following Vision-OPD, all token-level distribution computations use $K=100$ vocabulary entries.
At response position $t$, we first select the top-100 vocabulary entries according to the current full-image student distribution $p^{R}_{\theta,t}$.
The same vocabulary indices are then used to gather the corresponding logits from the crop-conditioned EMA teacher $p^{C}_{\bar{\theta},t}$ and the masked-image student $p^{R^{-}}_{\theta,t}$.
Therefore, the three distributions used by EAD are represented in the same vocabulary coordinates, which allows their distributional changes to be compared directly.
More explicitly, if
\[
\mathcal{S}_t
=
\operatorname{TopK}\!\left(p^{R}_{\theta,t}, K\right),
\qquad K=100,
\]
then all quantities in Eqs.~\ref{eq:ead_vectors}--\ref{eq:ead_weight} are computed after restricting
$p^{R}_{\theta,t}$,
$p^{C}_{\bar{\theta},t}$, and
$p^{R^{-}}_{\theta,t}$
to the common support $\mathcal{S}_t$.

\subsection{Preservation of broader visual capabilities.}
\label{app:generalization}
We additionally evaluate the held-out generalization on MMVP~\citep{tong2024eyes}, CV-Bench~\citep{tong2024cambrian}, MMStar~\citep{chen2024we}, and POPE~\citep{li2023evaluating}.
We use the same 4B and 9B model checkpoints as those evaluated in Table~\ref{tab:fine-grained}.
Table~\ref{tab:heldout} shows that EAD largely preserves the base model's average performance across the four held-out benchmarks, reducing the performance degradation relative to dense Vision-OPD by about 83\% at both model scales.
These results show that EAD improves fine-grained perception while largely preserving broader visual capabilities.

\begin{table}[t]
\centering
\small
\caption{
\textbf{Preservation of broader visual capabilities.}
We evaluate the same 4B and 9B post-training checkpoints from Table~\ref{tab:fine-grained} on four held-out benchmarks without
additional training. Avg.\ denotes the unweighted mean over the four benchmarks.
Best results within each model scale are shown in bold.
}
\label{tab:heldout}
\begin{tabular}{lccccc}
\toprule
\textbf{Method}
& \textbf{MMVP}
& \textbf{CV-Bench}
& \textbf{MMStar}
& \textbf{POPE}
& \textbf{Avg.} \\
\midrule

\multicolumn{6}{c}{
\textbf{Qwen3.5-4B Post-Training Strategies}
} \\
\midrule

Vanilla
& 79.00 & 86.97 & \textbf{74.00} & 89.33 & \textbf{82.33} \\

GRPO~\citep{shao2024deepseekmath}
& 78.00 & 85.78 & 69.33 & 87.87 & 80.25 \\

Vision-OPD~\citep{yuan2026vision}
& 76.67 & 85.48 & 70.80 & 88.99 & 80.49 \\

OPD-V~\citep{bi2026opd}
& 77.00 & 84.38 & 71.40 & \textbf{89.43} & 80.55 \\

VAD~\citep{zhang2026vad}
& 79.67 & \textbf{87.30} & 71.80 & 88.79 & 81.89 \\

\textbf{EAD (Ours)}
& \textbf{80.33} & 86.95 & 73.07 & 87.70 & 82.01 \\

\midrule
\multicolumn{6}{c}{
\textbf{Qwen3.5-9B Post-Training Strategies}
} \\
\midrule

Vanilla
& 82.67 & 88.41 & \textbf{77.87} & 89.86 & 84.70 \\

GRPO~\citep{shao2024deepseekmath}
& 80.33 & 87.87 & 72.80 & \textbf{89.97} & 82.74 \\

Vision-OPD~\citep{yuan2026vision}
& 81.00 & 86.52 & 74.33 & 89.80 & 82.91 \\

OPD-V~\citep{bi2026opd}
& 80.33 & 85.36 & 73.13 & 89.63 & 82.11 \\

VAD~\citep{zhang2026vad}
& \textbf{85.00} & \textbf{88.82} & 76.60 & 89.54 & \textbf{84.99} \\

\textbf{EAD (Ours)}
& 84.33 & 88.11 & 76.40 & 88.73 & 84.39 \\

\bottomrule
\end{tabular}
\end{table}

\subsection{Answer Extraction and Evaluation}
\label{app:evaluation}

\paragraph{Evaluation protocol.}
All Qwen3.5 post-training variants are evaluated under the same inference and evaluation pipeline.
To make multiple-choice grading more robust to free-form model responses, we revise the answer-extraction procedure in the released Vision-OPD evaluator and apply the revised evaluator uniformly to all variants.

\paragraph{Answer extraction.}
For multiple-choice responses, the released evaluator extracts the predicted option using a sequence of regular-expression patterns.
Its final fallback, \verb|([A-Z])|,  may extract an uppercase letter from ordinary response text rather than the selected option.
For example, for the response \texttt{Answer: **D**}, this fallback may match the \texttt{A} in \texttt{Answer} rather than the stated option \texttt{D}.

Therefore, we remove the unrestricted uppercase-letter fallback and only extract option labels from explicit answer expressions or standard option formats.
If no label can be resolved, the evaluator matches the response against the option texts and returns a prediction only when a single option is identified.
When multiple explicit answer expressions appear, the last one is taken as the final choice.
Only unresolved responses are passed to the fallback judge.

\paragraph{Fallback judging.}
For unresolved responses, we use Qwen3.5-122B-A10B~\citep{qwen35} as a text-only fallback judge.
The judge receives the question, reference answer, and generated response, but no image, and determines whether the response is equivalent to the reference.
We use temperature zero with thinking disabled and constrain the output to \texttt{Yes} or \texttt{No}.

\paragraph{Evaluation of the released Vision-OPD checkpoint.}
We evaluate the publicly released Vision-OPD 4B checkpoint with our revised evaluator.
This result serves as an additional reference for comparison under a consistent evaluation protocol.
\begin{table}[t]
\centering
\small
\setlength{\tabcolsep}{8pt}
\renewcommand{\arraystretch}{1.08}
\caption{
\textbf{Released Vision-OPD 4B checkpoint under our evaluation
pipeline.}
All results are obtained using the same revised evaluator applied to the Qwen3.5-based models in this work.
}
\label{tab:visionopd-official-4b}
\begin{tabular}{lc}
\toprule
\textbf{Benchmark} & \textbf{Vision-OPD 4B} \\
\midrule
V*Bench             & 90.05 \\
ZoomBench         & 59.29 \\
HRBench-4K        & 82.25 \\
HRBench-8K        & 79.25 \\
MME-RealWorld-EN  & 70.71 \\
MME-RealWorld-CN  & 69.43 \\
\midrule
MMVP              & 79.00 \\
CV-Bench          & 86.82 \\
MMStar            & 70.87 \\
POPE              & 89.12 \\
\midrule
\textbf{Fine Avg.}    & \textbf{75.16} \\
\textbf{Holdout Avg.} & \textbf{81.45} \\
\textbf{All-10}       & \textbf{77.68} \\
\bottomrule
\end{tabular}
\end{table}

\subsection{Details of the Supervision Retention Analysis}
\label{app:supervision-analysis}

\paragraph{Retained supervision mass.}
Figure~\ref{fig:supervision-analysis}(a) shows the fraction of distillation loss retained by EAD over the 65 training steps. Let $L_s^{\mathrm{dense}}$ and $L_s^{\mathrm{EAD}}$ denote the unweighted and EAD-weighted distillation losses at step $s$, respectively. The per-step and cumulative retained fractions are
\begin{equation}
R_s=\frac{L_s^{\mathrm{EAD}}}{L_s^{\mathrm{dense}}},
\qquad
R_{\le S}=
\frac{\sum_{s=1}^{S}L_s^{\mathrm{EAD}}}
     {\sum_{s=1}^{S}L_s^{\mathrm{dense}}}.
\end{equation}
Solid curves show $R_s$ and dashed curves show $R_{\le s}$.
The final cumulative fractions are $5.87\%$ for 4B and $6.15\%$ for 9B.
The dense reference is the unweighted loss on the same EAD training trajectories. 

\paragraph{Within-response supervision concentration.}
For Figure~\ref{fig:supervision-analysis}(b), we analyze 4B and 9B responses to the same 128 randomly sampled questions from training set.
Token-level losses and EAD weights are computed using the step-65 Student and EMA Teacher checkpoints.
For each response $i$, we sort the retained token losses $a_{i,t}=w_{i,t}D^{teacher}_{i,t}$ in descending order. With $n_i$ valid tokens, the fraction of total loss carried by the top $k$ tokens is
\begin{equation}
C_i\left(\frac{k}{n_i}\right)
=
\frac{\sum_{j=1}^{k}a_{i,(j)}}
     {\sum_{j=1}^{n_i}a_{i,(j)}},
\qquad k=0,\ldots,n_i,
\end{equation}
where $a_{i,(j)}$ is the $j$-th largest retained token loss.
The dense reference follows the same procedure, independently sorting the unweighted losses $D^{teacher}_{i,t}$.

\paragraph{On-policy rollout accuracy.}
Figure~\ref{fig:supervision-analysis}(c) compares 4B/9B EAD and Vision-OPD over 65 training steps. Each step uses 96 questions with eight responses per question. 
All responses are scored using the same answer extractor; unresolved answers count as incorrect. Faint curves show raw per-step accuracy. Bold curves show the mean accuracy over the current step and the preceding four steps.

\begin{table}[t]
\centering
\small
\caption{Summary of supervision retention, concentration, and rollout accuracy. All values are percentages.
Top-10\% loss mass measures how much of each response's total loss comes from its 10\% highest-loss
tokens, averaged across responses. 
Rollout accuracy is computed over all responses from training steps 1--65.
}
\label{tab:supervision-analysis-summary}
\begin{tabular}{lrr}
\toprule
Statistic & 4B & 9B \\
\midrule
Cumulative retained loss mass & 5.87 & 6.15 \\
Top-10\% loss mass: EAD & 95.98 & 96.81 \\
Top-10\% loss mass: dense & 61.10 & 62.27 \\
Training rollout accuracy: EAD & 59.53 & 64.31 \\
Training rollout accuracy: Vision-OPD & 55.06 & 57.49 \\
\bottomrule
\end{tabular}
\end{table}

\subsection{Held-Out Generalization of Evidence References}
\label{app:evidence-reference-generalization}

We further evaluate the evidence-reference variants from Section~\ref{sec:evidence-reference-ablation} on four held-out benchmarks using the same trained Qwen3.5-4B checkpoints.
Table~\ref{tab:evidence-reference-generalization} reports the held-out results together with the Fine Avg.\ from the six fine-grained benchmarks and the average over all ten benchmarks.

\begin{table}[t]
\centering
\small
\setlength{\tabcolsep}{5.5pt}
\renewcommand{\arraystretch}{1.08}
\caption{
\textbf{Held-out generalization of evidence-reference variants.}
The same checkpoints evaluated in
Tables~\ref{tab:evidence-model} and~\ref{tab:evidence-visual}
are evaluated on MMVP, CV-Bench, MMStar, and POPE.
Holdout Avg.\ is the unweighted mean over the four held-out
benchmarks, Fine Avg.\ is the mean over the six fine-grained
benchmarks, and All-10 averages all ten benchmarks.
}
\label{tab:evidence-reference-generalization}
\resizebox{\textwidth}{!}{
\begin{tabular}{lccccccc}
\toprule
\textbf{Evidence Reference}
& \textbf{MMVP}
& \textbf{CV-Bench}
& \textbf{MMStar}
& \textbf{POPE}
& \textbf{Holdout Avg.}
& \textbf{Fine Avg.}
& \textbf{All-10} \\
\midrule

Cross-state masking
& 77.00
& 85.49
& \textbf{73.53}
& 86.66
& 80.67
& 75.98
& 77.86 \\

Crop-induced change
& 78.33
& 86.61
& 70.87
& 89.11
& 81.23
& 75.16
& 77.59 \\

Teacher-side masking
& 78.67
& \textbf{87.59}
& 72.67
& \textbf{89.24}
& \textbf{82.04}
& 77.03
& 79.03 \\

Teacher crop degradation
& 76.33
& 84.15
& 71.87
& 83.30
& 78.91
& 76.33
& 77.37 \\

Current student (EAD)
& \textbf{80.33}
& 86.95
& 73.07
& 87.70
& 82.01
& \textbf{77.53}
& \textbf{79.33} \\

\bottomrule
\end{tabular}
}
\end{table}

The held-out results are consistent with the main fine-grained comparison.
Both same-state Original--Mask variants outperform the cross-state construction, supporting the benefit of removing teacher--student state differences from the evidence reference.
Among the visual comparisons, Original--Mask also generalizes better than the teacher-side crop-degradation reference.
Although teacher-side masking achieves a marginally higher Holdout Avg.\ than EAD, EAD attains the strongest Fine Avg.\ and All-10 score while maintaining comparable held-out performance.

\subsection{Effect of Local JS Scaling}
\label{app:local-js-scaling}

EAD represents both the privileged teacher correction and the
learner evidence response in the locally normalized coordinates
motivated by the second-order approximation of Jensen--Shannon
divergence.
We ablate this design by directly computing cosine similarity
between the corresponding probability differences without the
$1/\sqrt{p_{\theta,t}^{R}}$ scaling.

Specifically, the unscaled variant uses
\begin{equation}
\tilde{u}^{\mathrm{corr}}_t
=
p^{C}_{\bar{\theta},t}
-
p^{R}_{\theta,t},
\qquad
\tilde{u}^{\mathrm{evid}}_t
=
p^{R}_{\theta,t}
-
p^{R^-}_{\theta,t},
\end{equation}
and computes
\begin{equation}
\tilde{w}_t
=
\max\left(
\cos\left(
\tilde{u}^{\mathrm{corr}}_t,
\tilde{u}^{\mathrm{evid}}_t
\right),
0
\right).
\end{equation}
All other training settings are kept identical to EAD.

\begin{table}[t]
\centering
\small
\setlength{\tabcolsep}{8pt}
\renewcommand{\arraystretch}{1.08}
\caption{
\textbf{Effect of local JS scaling.}
Vanilla cosine directly compares the unscaled probability
differences, whereas EAD applies the shared local normalization
$1/\sqrt{\max(p_{\theta,t}^{R},\epsilon)}$ before measuring directional alignment.
Both variants otherwise use the same evidence reference and
training procedure.
}
\label{tab:local-js-scaling}
\begin{tabular}{lcc}
\toprule
\textbf{Benchmark} &
\textbf{Vanilla Cosine} &
\textbf{EAD} \\
\midrule
V*Bench              & 89.01 & \textbf{91.62} \\
ZoomBench          & 59.29 & \textbf{60.83} \\
HRBench-4K         & 85.12 & \textbf{86.38} \\
HRBench-8K         & 82.38 & \textbf{82.88} \\
MME-RealWorld-EN   & 72.37 & \textbf{72.39} \\
MME-RealWorld-CN   & \textbf{71.29} & 71.10 \\
\midrule
MMVP                & 79.67 & \textbf{80.33} \\
CV-Bench            & \textbf{87.03} & 86.95 \\
MMStar              & \textbf{73.47} & 73.07 \\
POPE                & 87.57 & \textbf{87.70} \\
\midrule
\textbf{Fine Avg.}    & 76.58 & \textbf{77.53} \\
\textbf{Holdout Avg.} & 81.94 & \textbf{82.01} \\
\textbf{All-10}       & 78.72 & \textbf{79.33} \\
\bottomrule
\end{tabular}
\end{table}

As shown in Table~\ref{tab:local-js-scaling}, directly comparing the original probability differences already provides a meaningful alignment signal, while local normalization yields stronger performance.
These results support the use of the shared local JS geometry.

\subsection{Effect of Excluding Negatively Aligned Corrections}
\label{app:negative-alignment}

EAD uses positive cosine weighting, so teacher corrections with negative alignment to the evidence reference are excluded from supervision.
We test this choice with a shifted cosine that retains negatively aligned corrections.
Specifically, the shifted-cosine variant uses
\begin{equation}
\tilde{w}_t
=
\frac{
\cos\left(
u_t^{\mathrm{corr}},
u_t^{\mathrm{evid}}
\right)+1
}{2},
\end{equation}
while keeping all other training settings unchanged.
\begin{table}[t]
\centering
\small
\setlength{\tabcolsep}{8pt}
\renewcommand{\arraystretch}{1.08}
\caption{
\textbf{Effect of excluding negatively aligned corrections.}
Shifted-Cosine retains negatively aligned corrections by mapping cosine similarity from $[-1,1]$ to $[0,1]$.
All other settings are unchanged.
}
\label{tab:negative-alignment}
\begin{tabular}{lcc}
\toprule
\textbf{Benchmark}
& \textbf{Shifted-Cosine}
& \textbf{EAD} \\
\midrule
V*Bench              & 90.58 & \textbf{91.62} \\
ZoomBench          & \textbf{61.18} & 60.83 \\
HRBench-4K         & 83.50 & \textbf{86.38} \\
HRBench-8K         & 81.25 & \textbf{82.88} \\
MME-RealWorld-EN   & 72.22 & \textbf{72.39} \\
MME-RealWorld-CN   & 70.05 & \textbf{71.10} \\
\midrule
MMVP                & 77.33 & \textbf{80.33} \\
CV-Bench            & \textbf{87.21} & 86.95 \\
MMStar              & \textbf{73.93} & 73.07 \\
POPE                & \textbf{88.88} & 87.70 \\
\midrule
\textbf{Fine Avg.}    & 76.46 & \textbf{77.53} \\
\textbf{Holdout Avg.} & 81.84 & \textbf{82.01} \\
\textbf{All-10}       & 78.61 & \textbf{79.33} \\
\bottomrule
\end{tabular}
\end{table}
As shown in Table~\ref{tab:negative-alignment}, Shifted-Cosine yields lower overall performance than EAD.
These results support excluding opposing teacher corrections from supervision.

\begin{figure*}[t]
    \centering
    \includegraphics[width=\linewidth]{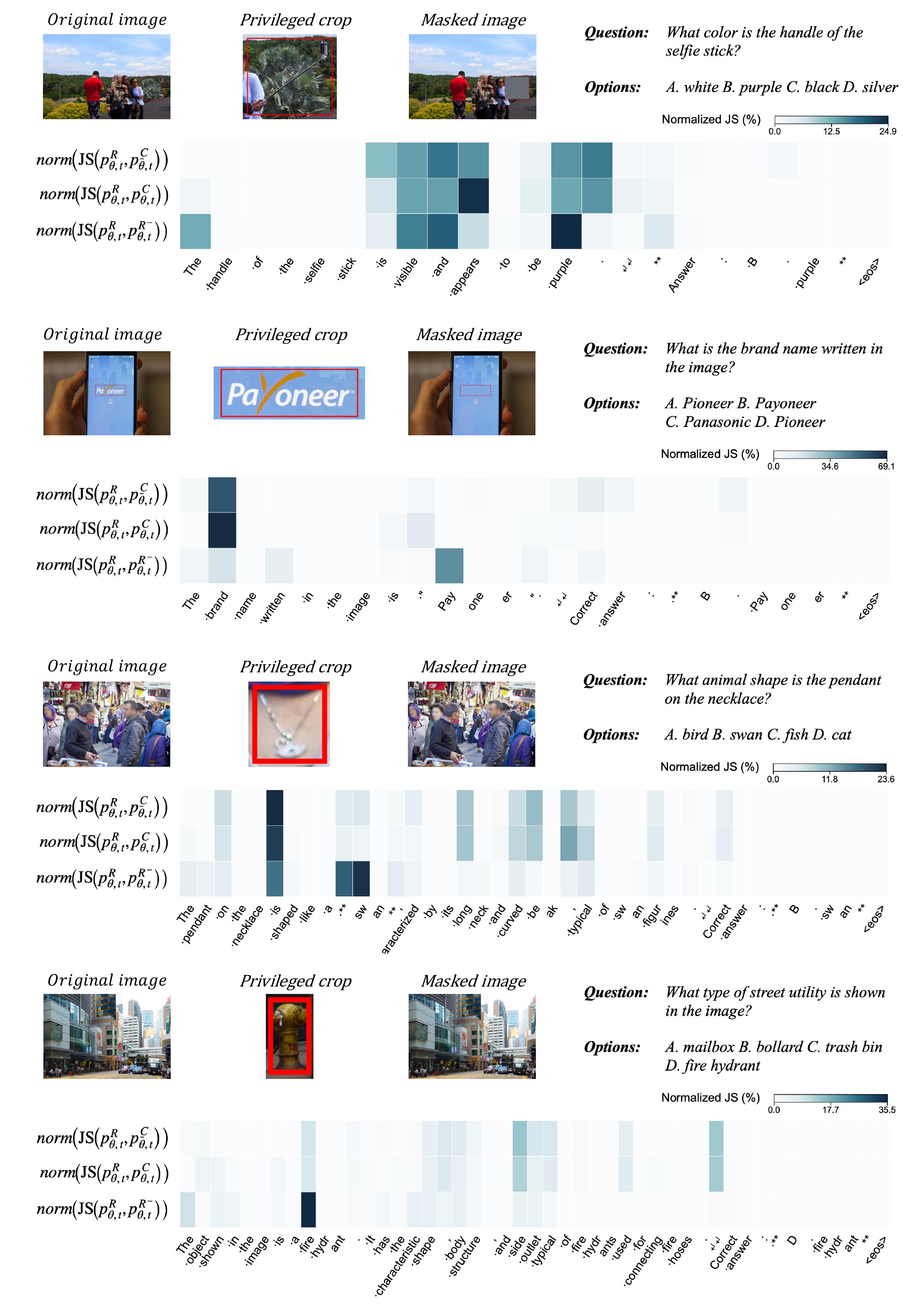}
    \caption{
    \textbf{Additional token-level case studies.}
    Each example shows the original, cropped, and masked images together with the normalized token-wise JS divergence of $D_t^{\mathrm{teacher}}$, $D_t^{\mathrm{crop}}$, and $D_t^{\mathrm{mask}}$.
    Across different fine-grained visual tasks, $D_t^{\mathrm{mask}}$ tends to emphasize tokens more closely associated with the task-relevant visual evidence, while the other comparisons can assign large divergence to less directly related tokens.
    }
    \label{fig:additional-cases}
\end{figure*}

\subsection{Additional Token-Level Case Studies}
\label{app:case-studies}

Figure~\ref{fig:additional-cases} provides additional examples across different fine-grained visual tasks, including attribute recognition, text recognition, shape recognition, and object identification.
For each example, we visualize the normalized token-wise JS divergence of $D_t^{\mathrm{teacher}}$, $D_t^{\mathrm{crop}}$, and $D_t^{\mathrm{mask}}$ on the same generated response.

Across these cases, $D_t^{\mathrm{mask}}$ places greater emphasis on tokens associated with the task-relevant visual evidence, whereas $D_t^{\mathrm{crop}}$ and $D_t^{\mathrm{teacher}}$  exhibit large divergence on tokens less directly related to that evidence.
These examples are consistent with the observation in Figure~\ref{fig:profiles} and further motivate using the Original--Mask comparison as the evidence reference.

\end{document}